\documentclass[conference]{IEEEtran}
\IEEEoverridecommandlockouts
\usepackage{pgfplots}
\pgfplotsset{compat=1.18} 
\usepackage{cite}
\usepackage{enumitem}
\usepackage{calc}
\usepackage{amsmath,amssymb,amsfonts}
\usepackage{algorithmic}
\usepackage{graphicx}
\usepackage{textcomp}
\usepackage{xcolor}
\usepackage{tikz}
\usepackage{booktabs}
\usepackage[absolute,overlay]{textpos}
\usepackage{lipsum}                  
\usetikzlibrary{arrows,shapes,positioning}
\usepackage{afterpage}
\usepackage{float}
\usepackage{tabularx} 
\usepackage{booktabs}
\usepackage{siunitx} 
\def\BibTeX{{\rm B\kern-.05em{\sc i\kern-.025em b}\kern-.08em
    T\kern-.1667em\lower.7ex\hbox{E}\kern-.125emX}}
\usepackage{tikz}

\newcommand\copyrighttext{%
  \footnotesize \textcopyright 2025 IEEE. Personal use of this material is permitted.
  Permission from IEEE must be obtained for all other uses, in any current or future
  media, including reprinting/republishing this material for advertising or promotional
  purposes, creating new collective works, for resale or redistribution to servers or
  lists, or reuse of any copyrighted component of this work in other works.}
\newcommand\copyrightnotice{%
\begin{tikzpicture}[remember picture,overlay]
\node[anchor=south,yshift=10pt] at (current page.south) 
  {\fbox{\parbox{\dimexpr\textwidth-\fboxsep-\fboxrule\relax}{\copyrighttext}}};
\end{tikzpicture}%
}

\begin{document}

\title{Hi-DARTS: Hierarchical Dynamically Adapting Reinforcement Trading System\\}

\author{
    \IEEEauthorblockN{HOON SAGONG}
    \IEEEauthorblockA{\textit{Dept. of Computer Engineering} \\
    \textit{Hongik University}\\
    Seoul, South Korea \\
    clap518@mail.hongik.ac.kr}
\and
    \IEEEauthorblockN{HEESU KIM}
    \IEEEauthorblockA{\textit{Dept. of Applied Data Science} \\
    \textit{Sungkyunkwan University}\\
    Seoul, South Korea \\
    k.heesu@skku.edu}
\and
    \IEEEauthorblockN{HANBEEN HONG}
    \IEEEauthorblockA{\textit{Dept. of Economics} \\
    \textit{Hankuk University of Foreign Studies}\\
    Seoul, South Korea \\
    hhb0618@hufs.ac.kr}
}

\maketitle
\copyrightnotice

\begin{abstract}
Conventional autonomous trading systems struggle to balance computational efficiency and market responsiveness due to their fixed operating frequency. We propose Hi-DARTS, a hierarchical multi-agent reinforcement learning framework that addresses this trade-off. Hi-DARTS utilizes a meta-agent to analyze market volatility and dynamically activate specialized Time Frame Agents for high-frequency or low-frequency trading as needed. During back-testing on AAPL stock from January 2024 to May 2025, Hi-DARTS yielded a cumulative return of 25.17\% with a Sharpe Ratio of 0.75. This performance surpasses standard benchmarks, including a passive buy-and-hold strategy on AAPL (12.19\% return) and the S\&P 500 ETF (SPY) (20.01\% return). Our work demonstrates that dynamic, hierarchical agents can achieve superior risk-adjusted returns while maintaining high computational efficiency.

\end{abstract}


\section{Introduction}
The proliferation of high-frequency trading in financial markets has made automated trading systems essential for achieving competitive yields. However, these systems struggle with balancing computational efficiency with market responsiveness. Most frameworks active today operate at a fixed frequency, consuming computational resources in a consistent manner. This is a static approach that is either computationally wasteful during calm market periods or too slow to capitalize on opportunities during periods of high volatility. This tradeoff between cost and performance creates a demand for adaptive solutions. 
Recent advances in deep reinforcement learning have shown promising results in high-performance decision-making problems\cite{b1}. Multi-agent reinforcement learning frameworks have been found to perform effectively for complex tasks\cite{b2,b3,b4}. 
To address this challenge, we introduce Hi-DARTS, a novel Hierarchical Dynamically Adapting Reinforcement Trading System. Our framework utilizes a two-layer structure of Proximal Policy Optimization (PPO) agents. The top-level agent, the Time Frame Allocator, analyzes real-time market data and employs one of several specialized Time Frame Agents, each trained to operate optimally at a different frequency (e.g. 1 hour, 10 minute, or 1 minute). This hierarchical design allows the system to conserve resources in stable markets and remain capable of reacting to fast changes. 

The main contributions of this paper are threefold:
\begin{itemize}
    \item We propose a novel hierarchical multi-agent architecture for algorithmic trading that dynamically adapts its operational frequency to market conditions.
    \item We introduce a dynamic allocation mechanism where a meta-agent effectively learns to select the optimal specialized trading agent based on volatility.
    \item We provide empirical validation of our framework on real-world stock data, demonstrating that Hi-DARTS achieves superior risk-adjusted returns compared to static benchmarks.
\end{itemize}

\section{Proposed Hierarchical Framework}
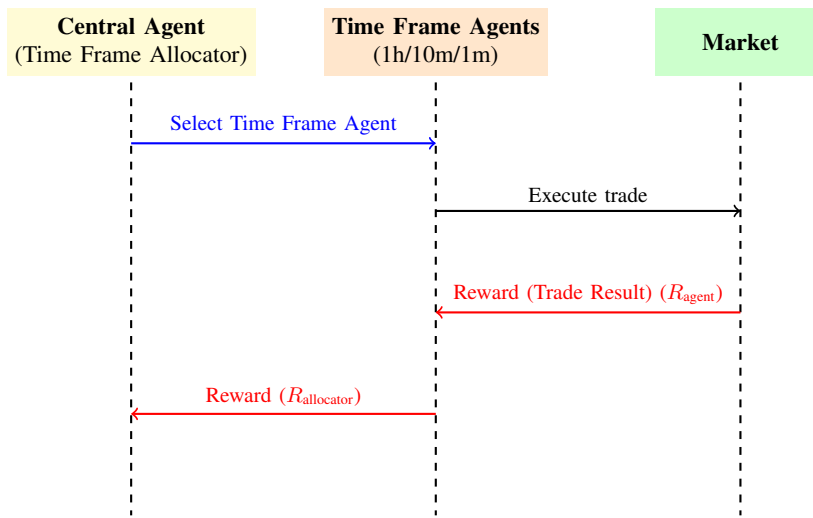
\begin{figure*}[!t]
    \centering
    \begin{tikzpicture}[scale=0.9, every node/.style={transform shape}]
        \def\centralx{0}
        \def\agentx{4.5}
        \def\marketx{9}
        
        \node[rectangle, fill=yellow!20, minimum width=2.5cm, minimum height=1cm, align=center] at (\centralx, 0) {\textbf{Central Agent} \\ (Time Frame Allocator)};
        \node[rectangle, fill=orange!20, minimum width=2.5cm, minimum height=1cm, align=center] at (\agentx, 0) {\textbf{Time Frame Agents} \\ (1h/10m/1m)};
        \node[rectangle, fill=green!20, minimum width=2.5cm, minimum height=1cm, align=center] at (\marketx, 0) {\textbf{Market}};
        
        \draw[thick, dashed] (\centralx, -0.6) -- (\centralx, -7);
        \draw[thick, dashed] (\agentx, -0.6) -- (\agentx, -7);
        \draw[thick, dashed] (\marketx, -0.6) -- (\marketx, -7);
        
        \draw[->, thick, blue] (\centralx, -1.5) -- (\agentx, -1.5);
        \node[above, blue, font=\small] at (2.25, -1.5) {Select Time Frame Agent};
        
        \draw[->, thick, black] (\agentx, -2.5) -- (\marketx, -2.5);
        \node[above, black, font=\small] at (6.75, -2.5) {Execute trade};
        
        \draw[->, thick, red] (\marketx, -4) -- (\agentx, -4);
        \node[above, red, font=\small] at (6.75, -4) {Reward (Trade Result) ($R_{\text{agent}}$)};
        
        \draw[->, thick, red] (\agentx, -5.5) -- (\centralx, -5.5);
        \node[above, red, font=\small] at (2.25, -5.5) {Reward ($R_{\text{allocator}}$)};
        
    \end{tikzpicture}
    \caption{Reward flow in the adaptive framework: The trade outcome from the Market provides a direct reward ($R_{\text{agent}}$) to the selected Time Frame Agent for its own learning. This outcome is then used to generate a subsequent reward ($R_{\text{allocator}}$) for the Central Agent, allowing it to learn the optimal allocation strategy.}
    \label{fig:reward_flow_architecture}
\end{figure*}

The proposed framework consists of a two-layered hierarchical structure, as illustrated in Fig.~\ref{fig:reward_flow_architecture}. This top-down design is a modular approach that has numerous benefits in this particular system, such as the specialization of each task, selecting the Time Frame Agent and decision making. This specialization allows the system to perform better than the traditional layerless trading system\cite{b5}. Then, by using a risk-based reward system with each layer, the system has the ability to evaluate its own performance and further strengthen its capabilities. With the help of propagated rewards, the Time Frame Allocator can select the appropriate Time Frame Agents.

\subsection{Stock Layer (Time Frame Allocator)}
This layer is the adaptive mechanism and functions as a Time Frame Agents Allocator. It continuously analyzes the selected stock's recent market data with the rewards produced from the Time Frame Agents, to assess the current market state of the stock. Based on this assessment, it determines which of the Time Frame Agents is most suitable for the current conditions and should be designated. For example, in a highly volatile market, it would activate a high-frequency agent (e.g., 1 minute agent), whereas in a stable market, it would initiate a low-frequency agent (e.g., 1 hour agent) to conserve resources.

\subsection{Time Frame Agents (Market Responders)}
At the lowest level, the Time Frame Agents directly interact with the raw stock market data and come up with the judgment to buy, sell, or hold. These agents are also tasked with making the transaction itself, acting semi-autonomously since they are directly managed by the upper layer. The agents are trained independently, using only the data with the same time resolution. By limiting the data these agents can view, they are able to better analyze the given data, since they are more specialized to deal with data with such time resolution. When deployed, they are each fed with relevant data, and each of them comes up with their own judgment of the situation by reacting to the real-time data. Conclusively, after making such actions, rewards are produced from the outcome of their actions.

\subsubsection{Technical Indicators and Variables}
Table~\ref{tab1} contains the features selected for our models. These technical indicators and portfolio statistics serve as input variables for each Time Frame Agent to inform trading decisions at their respective operational frequency.
The selected features are as follows:
\begin{itemize}
    \item RSI 14: The 14-period Relative Strength Index, a momentum oscillator that measures the speed and change of price movements, helping to identify overbought or oversold conditions.
    \item MACD Hist: The Moving Average Convergence Divergence Histogram, which captures the difference between the MACD line and its signal line, providing insight into the strength and direction of momentum.
    \item CCI20: The 20-period Commodity Channel Index, used to identify cyclical trends in security prices and detect deviation from typical price ranges.
    \item BB pband 20: The 20-period Bollinger Bands \%B, describing the position of the latest closing price with respect to the Bollinger Bands, indicating potential breakouts or mean-reversion opportunities.
    \item Volume: The total number of shares traded over the specified period, reflecting market activity and liquidity.
    \item Cash Ratio: The proportion of the agent’s portfolio held in cash, calculated as the cash balance divided by the total portfolio value. This ratio provides information on available liquidity and risk management posture.
    \item Stock Ratio: The proportion of the agent’s portfolio invested in equities, computed as the stock value divided by the total portfolio value, indicating market exposure.
    \item Unrealized Profit Ratio: The percentage profit or loss of currently held equity positions, which lets the agent to assess ongoing trade performance and adjust actions accordingly.
\end{itemize}

These indicators collectively enable the Time Frame Agents to capture momentum, volatility, trend, market activity, and portfolio health, supporting robust decision-making across different market regimes.

\begin{table}[t]
    \small
    \centering
    \caption{Selected Features and Descriptions} 
    \label{tab:features}
    \begin{tabularx}{\columnwidth}{@{}l X@{}}
    \toprule
    \textbf{Features}                & \textbf{Descriptions} \\
    \midrule  
    RSI\_14                     & 14-period Relative Strength Index \\
    MACD\_Hist                  & Moving Average Convergence Divergence histogram \\
    CCI20                       & 20-period Commodity Channel Index \\
    BB\_pband\_20               & 20-period Bollinger Bands \%B \\
    Volume                      & Total number of shares traded over a period \\
    Cash\_Ratio                 & Cash balance divided by total portfolio value \\
    Stock\_Ratio                & Stock value divided by total portfolio value \\
    Unrealized\_Profit\_Ratio   & The profit or loss in percentage of the equity held \\
    \bottomrule
    \end{tabularx}
    \label{tab1}
\end{table}

\section{Implementation and Validation}
We implemented our proposed framework using Proximal Policy Optimization (PPO) as the base algorithm for our agents. PPO was chosen because it has shown superior performance over policy gradient methods\cite{b6}. All experiments were conducted on a single GPU (Tesla V100). All data used to train, validate, and test came from the Wharton Research Data Services (WRDS). To validate this implementation, we designed benchmarks that demonstrate the ability of the allocator to make the optimal decision when given the data. Consequently, Time Frame Agents’ (1 minute, 10 minute, 1 hour) and Central Agent’s (Time Frame Allocator) architecture is made of two hidden layer MLP for policy and value networks. Key hyperparameters, such as learning rate, state window size, and total timesteps, were individually tuned for each agent's respective timeframe to optimize its performance. All hyperparameters for each agent is depicted in Table~\ref{hyperparameters}.
\subsection{Experimental Setup}
In the experimental setup for this proof-of-concept implementation, our system was trained and evaluated on Apple Inc. (AAPL) stock data. Three different Time Frame Agents were trained on 1 minute, 10 minute, and 1 hour interval data. The dataset was divided into three parts: training, validation, and testing. The training set consists of data from January 2, 2013, to December 31, 2022, the validation set from January 2, 2023, to December 31, 2023, and the test set from January 2, 2024, to May 2, 2025. Each Time Frame Agent was trained for 200,000 timesteps in total, with an initial cash budget of \$10,000. The Stock Layer (Time Frame Allocator) was then trained on the 1 minute data from the trainset with pre-trained Time Frame Agents attached.
\begin{table}[b!]
\centering
\caption{Detailed hyperparameters for all agents used in the Hi-DARTS framework.}
\label{tab:hyperparameters}
\begin{tabular}{l|c|c|c|c}
\hline
\textbf{Hyperparameter} & \textbf{Allocator} & \textbf{1m Agent} & \textbf{10m Agent} & \textbf{1h Agent} \\
\hline
\multicolumn{5}{c}{\textit{PPO Algorithm Parameters}} \\
\hline
Total Timesteps     & 300,000                & 500,000           & 200,000             & 150,000            \\
Learning Rate       & 3e-4                   & 5e-5              & 1e-4                & 1e-4               \\
N\_Steps            & 2048                   & 4096              & 2048                & 1024               \\
Batch Size          & 256                    & 128               & 128                 & 64                 \\
N\_Epochs           & 10                     & 10                & 10                  & 10                 \\
Gamma   & 0.99                   & 0.99              & 0.99                & 0.99               \\
GAE Lambda & 0.95                & 0.95              & 0.95                & 0.95               \\
Clip Range          & 0.2                    & 0.2               & 0.2                 & 0.2                \\
Entropy Coef. & 0.005             & 0.01              & 0.03                & 0.01               \\
\hline
\multicolumn{5}{c}{\textit{Environment \& State Parameters}} \\
\hline
Window Size   & N/A                    & 240               & 120                 & 80                 \\
Initial Cash        & \$10,000               & \$10,000          & \$10,000            & \$10,000           \\
\hline
\multicolumn{5}{c}{\textit{Network Architecture}} \\
\hline
MLP Layers & 2x64      & 2x256         & 2x64            & 2x256          \\
\hline
\end{tabular}
\label{hyperparameters}
\end{table}

\subsection{Performance Metrics}
To quantitatively validate our framework, we tested its performance against two standard benchmarks: a Straightforward Buy \& Hold strategy for the AAPL stock, and the S\&P500 Index ETF. Our evaluation focused on the following metrics:
\begin{itemize}
    \item Cumulative Return: The total gain over a given period of time.
    \item Sharpe Ratio: Risk-adjusted performance compared to a risk-free investment.
    \item Maximum Drawdown (MDD): Measure of the largest drop in the portfolios' value within the entire evaluation period. 
\end{itemize}
The Sharpe Ratio measures the performance of an investment by its risk-adjusted return as described in\cite{b7}. Maximum Drawdown, a standard risk metric in quantitative finance, measures the peak-to-trough decline of the investment for a given time period, as described in\cite{b8}.
\subsection{Reward Function}
Hi-DARTS uses the following reward functions for the Time Frame Agents and the Time Frame Allocator. The reward function is designed to reflect how profitable an executed trade is. Equation \eqref{Ragent} is defined as a hyperbolic tangent transformation of the normalized difference between the selling price and the average buying price of the stocks,
\begin{equation}
\label{Ragent}
    R_{\text{agent}} = \tanh \left( 5 \times \frac{P_{\text{sell}} - P_{\text{buy,avg}}}{P_{\text{buy,avg}}} \right)
\end{equation}
where $P_{sell}$ is the selling price, and $P_{buy,avg}$ is the average purchase price of the stock held. This formula ensures that the reward remains bounded between -1 and 1, providing stable gradients for training while emphasizing the relative profit made by each transaction.

The Time Frame Allocator, which dynamically selects among the specialized agents, receives feedback based on the overall portfolio value changes. Its reward $R_{allocator}$ is defined in \eqref{Ralloc} as the natural logarithm of the ratio between the current portfolio value $V_{current}$ and the previous portfolio value $V_{previous}$:
\begin{equation}
\label{Ralloc}
    R_\text{allocator} = \ln \left( \frac{V_{\text{current}}}{V_{\text{previous}}} \right)
\end{equation}
This logarithmic return metric captures the agent's ability to allocate trading decisions in a manner that maximizes portfolio growth while implicitly penalizing losses, facilitating the learning of an optimal allocation strategy that adapts to market dynamics.

By adopting these reward functions, Hi-DARTS effectively balances the micro-level trade execution performance of individual Time Frame Agents with the macro-level portfolio management objectives of the allocator, leading to improved risk-adjusted returns and adaptive trading behavior.
In order to have an immediate response, the 1 minute agent is selected by default at the start of the market. This helps the system react spontaneously and produce rewards as soon as the market opens. Also, the system is designed to liquidate all stock holdings by market close, aiming to optimize outcomes and constructively enhance agent performance.

\subsection{Stock Layer Validation}
Historical data from January 2, 2013 to December 31, 2022 was used to train the Time Frame Agents. This period contains both volatile and stable periods that taught the models to react to both types of markets. The allocators decide upon which Time Frame Agent to allocate by using previous data and reward feedback. The system was given data from January 2, 2021 to December 31, 2023 for the validation of the entire model. The allocator successfully opted for shorter frequency agents during periods of turbulent price movements and switched to prolonged frequency agents when the market was calm. This result confirms that our adaptive allocation mechanism produces the intended outcome, laying a solid foundation for the full system.

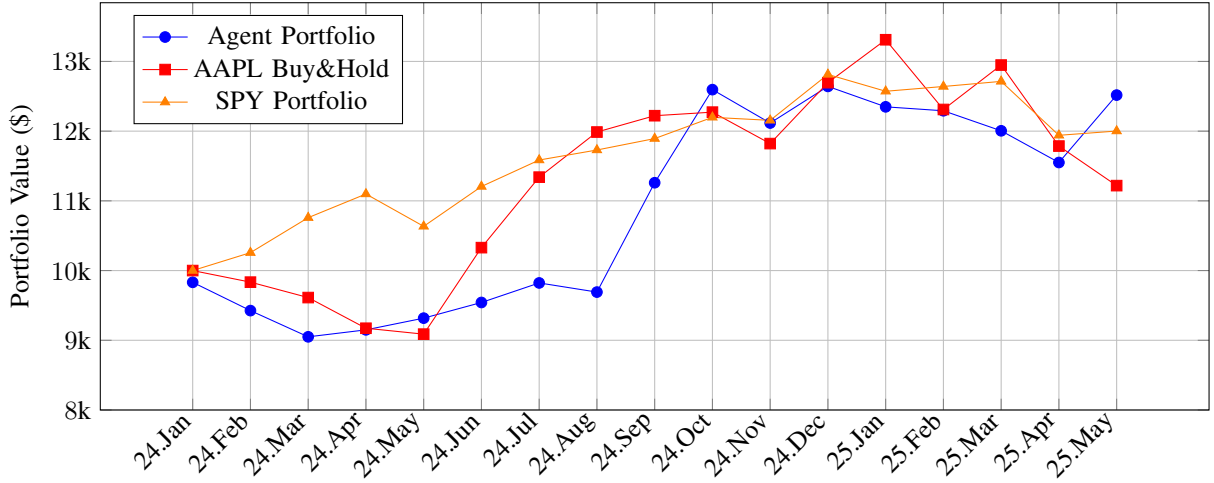
\begin{figure*}[t!] 
    \centering
    \begin{tikzpicture}
        \pgfplotstableread[col sep=comma]{
            Month,      Agent_Portfolio,    AAPL_BuyHold,   SPY_Portfolio
            24.Jan,     9831.37,            10000.00,       10000.00
            24.Feb,     9425.60,            9833.93,        10257.02
            24.Mar,     9049.57,            9612.62,        10758.89
            24.Apr,     9149.48,            9172.70,        11098.51
            24.May,     9317.52,            9087.98,        10634.36
            24.Jun,     9542.21,            10328.90,       11205.81
            24.Jul,     9822.53,            11340.08,       11585.62
            24.Aug,     9691.30,            11987.16,       11729.56
            24.Sep,     11259.91,           12221.36,       11893.42
            24.Oct,     12595.40,           12272.53,       12198.10
            24.Nov,     12116.40,           11821.48,       12154.82
            24.Dec,     12641.76,           12695.15,       12813.31
            25.Jan,     12348.60,           13310.94,       12572.04
            25.Feb,     12292.17,           12310.67,       12640.51
            25.Mar,     12004.84,           12947.90,       12713.79
            25.Apr,     11550.18,           11785.81,       11940.29
            25.May,     12517.04,           11218.51,       12001.43
       
        }\mydata
        \begin{axis}[
            width=0.9\textwidth, 
            height=7cm,          
            title={Portfolio Performance Comparison},
            xlabel={},
            ylabel={Portfolio Value (\$)},
            symbolic x coords={24.Jan, 24.Feb, 24.Mar, 24.Apr, 24.May, 24.Jun, 24.Jul, 24.Aug, 24.Sep, 24.Oct, 24.Nov, 24.Dec, 25.Jan, 25.Feb, 25.Mar, 25.Apr, 25.May},
            xtick=data,
            xticklabel style={rotate=45, anchor=east},
            legend pos=north west,
            grid=major,
            ymin=8000,
            scaled y ticks=false,
            yticklabel={%
                \pgfmathparse{\tick/1000}%
                \pgfmathprintnumber{\pgfmathresult}k%
            },
        ]
        \addplot[blue, mark=*] table[x=Month, y=Agent_Portfolio] {\mydata};
        \addlegendentry{Agent Portfolio}
        \addplot[red, mark=square*] table[x=Month, y=AAPL_BuyHold] {\mydata};
        \addlegendentry{AAPL Buy\&Hold}
        \addplot[orange, mark=triangle*] table[x=Month, y=SPY_Portfolio] {\mydata};
        \addlegendentry{SPY Portfolio}
        \end{axis}
    \end{tikzpicture}
    \caption{Comparison of the performance of Hi-DARTS and passive buy-and-hold strategies for AAPL and SPY during the test set time period. The test period is defined to be from January 2, 2024 to May 2, 2025. Hi-DARTS achieved a cumulative return of 25.17\% compared to the 12.19\% and 20.01\% of AAPL and SPY, respectively.}
    \label{fig:portfolio_performance_wide}
\end{figure*}
\begin{figure}[t]
\centerline{\includegraphics[width=\columnwidth]{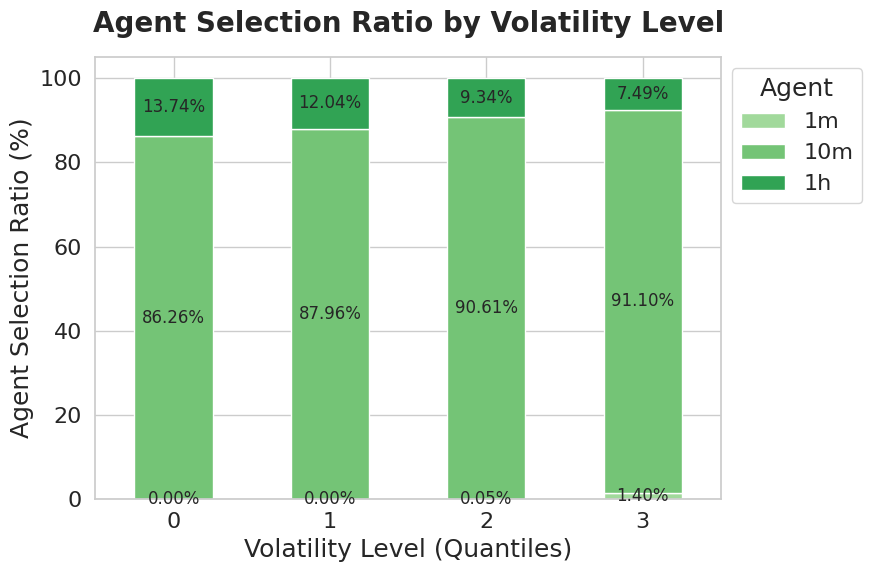}} 
\caption{Agent selection ratios are divided by daily volatility. The daily volatility of the market was calculated and then sorted into four equal quartiles, from lowest (0) to highest (3). This chart shows the average agent selection ratio for all days within each volatility quartile, confirming that the system activates high-frequency agents during volatile periods and low-frequency agents during stable periods.}
\label{fig_results}
\end{figure}

To analyze the designation of the agents between 1 minute, 10 minute, and 1 hour agents, we have analyzed the selection of agents monthly, daily, and hourly. To examine the allocation, the whole trade has been divided into quartiles, from the lowest volatility to the highest, as shown in Fig.~\ref{fig_results}. For this analysis, daily volatility was calculated as the standard deviation of price returns multiplied by 100.

Starting with the monthly report of agents allocation, the lowest quartile segment has shown the 0.18\% average pick rate for the 1 minute agent, while the average pick rate for 1 hour agent was 11.03\%. Then the second segment exhibited the 0.25\% utilization for 1 minute agent and 11.57\% usage of the 1 hour agent. For the third quartile, the 1 minute agent was selected 0.40\%, 1 hour agent was selected 10.43\%. Lastly, in the most volatile quartile, the pick rate for 1 minute has increased to 0.71\%, meanwhile, 1 hour rate was decreased to 7.90\%. 

Subsequently, the same trend continued in the daily average volatility, too. The first quartile had not employed the 1 minute agent at all, while the system used 1 hour agent 13.74\% of the time. In a similar manner, the second quartile also had not selected 1 minute agent for trading, whereas it had chosen 1 hour agent 12.04\%. Next, the third quartile deployed 1 minute agent 0.05\%, whilst 1 hour agent was designated 9.34\%. Lastly, the most versatile quartile had 1 minute agent 1.40\% of the time, though it used 1 hour agent 7.49\%. 

For the more granular examination, we have analyzed the hourly designation of the agents. Similarly to the analyzation above, the first quartile did not employ 1 minute agent at all, although it had employed 1 hour agent 18.54\%. Likewise, the second quartile had zero usage of 1 minute agent, but the system employed 1 hour agent 10.99\%. Correspondingly, the third quartile also did not use 1 minute agent; However, it deployed the 1 hour agent 6.95\%. The final quartile had 1.45\% of 1 minute agent and 6.19\% of 1 hour agent. 

With every month, day, hour agent selection, it clearly shows the trend; for the high volatility, the system opts for more 1 minute agent, subsequently fewer 1 hour agent. This has shown meaningful results from our stock layer, respectively, to the volatility it faces in the stock market.

\subsection{Result}
In this paper, we have introduced a hierarchical framework for an adaptive stock trading system. By dynamically allocating decision-making frequency based on ever-changing market conditions, our approach promises to balance computational efficiency with immediate market responsiveness. Our preliminary results validate the core component, the Time Frame Allocator, which successfully distinguishes different market states. Fees or taxes are intentionally excluded in our evaluation in order to generalize the measured performance across different environments, where transaction fees may vary. However, even with a standard transaction fee of 1 cent per sell, the final portfolio value's difference remains below 1\%.

Subsequently, the historical data from January 2, 2024 to May 2, 2025 was used to evaluate the system. The system was initially assigned with \$10,000.00, and the final portfolio was worth \$12,517.04, which produced the 25.17\% cumulative return (Table~\ref{tab2}). In comparison to the Buy \& Hold, which represents \$10,000 worth of stock purchased and held without being sold, the final valuation of that portfolio was \$11,218.51 during the same period, generating a 12.19\% rate of return (Table~\ref{tab2}). In addition, the S\&P rate of return during the same period was 20.01\%, making the S\&P500 portfolio worth \$12,001.43 (Table~\ref{tab2}). 

\section{Conclusion and Future Work}
Conventional automated trading systems are limited to fixed computing frequency, regardless of the market condition. It fails to adapt into a dynamic market. Consequently, if the performance of the automatic trading system is excessively set, it will waste resources in the calm markets. However, if minimized, it would not have enough responsiveness for the chaotic stock market.

Therefore, multi-agent frameworks can be more efficient in real-life situations. Hi-DARTS can save computational resources in the dynamic stock market. The product of this study validates the substantial correlation between the market's status and the agents allocation.

As demonstrated from the validation above, this system minimizes the waste of computational power compared to the traditional system. As the volatility level increases, the system adjusts the deployment ratio between the agents. In contrast, according to Fig.~\ref{fig_results}, at the first and second quartile, Hi-DARTS utilizes 0\% of 1 minute agent. This illustrates that at the low volatility levels, when the immediate response is not required, the system acts passively to save processing power. This implication is evidently exhibited by the Fig.~\ref{fig_results} bar graphs, which conveys Hi-DARTS making agent transitions in different volatility levels. 

As depicted from Fig.~\ref{fig_results}, when the market moved from the least volatile quartile to the most fluctuating quartile, the allocation of 1 minute agents was increased from 0\% to 1.4\%, while the 1 hour agent deployment decreased from 13.7\% to 7.5\%. When the market was volatile, the system made 1 minute agent more prominent to immediately respond to the fluctuations, meanwhile reducing the use of 1 hour agent to minimize the response time.

\begin{table}[t]
    \centering
    \footnotesize
    \setlength{\tabcolsep}{2pt}
    \caption{Performance Comparison with Benchmarks and Related Works}
    \label{tab:final_comparison_fixed2}
    \begin{tabular}{l c c c c}
        \toprule
        \textbf{Model} & \textbf{Test Period} & \textbf{Return (\%)} & \textbf{Sharpe Ratio} & \textbf{MDD (\%)} \\
        \midrule
        \textbf{Hi-DARTS} & \begin{tabular}{@{}l@{}}\textbf{2024.01--}\\\textbf{2025.05}\end{tabular} & \textbf{25.17} & \textbf{0.75} & \textbf{-25.49} \\
        1 Minute Agent & \begin{tabular}{@{}l@{}}2024.01--\\2025.05\end{tabular} & -8.06 & 0.04 & -25.78 \\
        10 Minute Agent & \begin{tabular}{@{}l@{}}2024.01--\\2025.05\end{tabular} & 10.16 & 0.35 & -18.68 \\
        1 Hour Agent & \begin{tabular}{@{}l@{}}2024.01--\\2025.05\end{tabular} & -1.38 & 0.05 & -21.67 \\
        \midrule
        Buy \& Hold (AAPL) & \begin{tabular}{@{}l@{}}2024.01--\\2025.05\end{tabular} & 12.19 & 0.35 & -33.36 \\
        S\&P 500 (SPY) & \begin{tabular}{@{}l@{}}2024.01--\\2025.05\end{tabular} & 20.01 & 0.58 & -18.76 \\
        \midrule
        \begin{tabular}{@{}l@{}}PPO (DJIA)\\(FinRL\cite{b9})\end{tabular} & \begin{tabular}{@{}l@{}}2019.01.01--\\2020.09.23\end{tabular} & 18.53 & 0.48 & -37.01 \\
        \bottomrule
    \end{tabular}
    \label{tab2}
\end{table}
Our future work will focus on completing the full suite of Time Frame Agents using PPO. We will also conduct a comprehensive performance evaluation by comparing our adaptive system's profitability and computational cost against the fixed-frequency automatic trading systems.

We plan to implement the Central Layer (Stock Selector) on top of the hierarchical model. The top layer of the hierarchy will be the Central Layer. Its primary responsibility is to analyze the stocks, depending on the existing data set. This layer will provide a specific stock to the lower layers. After finalizing all three layers of Hi-DARTS, we will apply other stocks to the system. The ultimate goal is to have the model seamlessly adapt to new assets without requiring additional training, due to its adaptive reward system.

Subsequent to finalizing the full suite, we plan to expand the number of Time Frame Agents to effectively cover the volatility of the perpetually shifting stock market. Then we plan to make an override mechanism that can overrule the initial designation of a low-frequency agent to high-frequency agent that is more reactive. This can assist the system to remain autonomous, even when market faces externalities that abruptly alter the market, such as the major geopolitical events or shocks that shake the stock market. 

Consequently, we have the objective of making a fully autonomous stock trading system that can save computational power while remaining responsive and effective.

\vspace{12pt}

\end{document}